\documentclass[conference]{IEEEtran}
\IEEEoverridecommandlockouts
\usepackage{booktabs}
\usepackage{amsmath,amssymb,amsfonts}
\usepackage{algorithmic}
\usepackage{graphicx}
\usepackage{textcomp}
\usepackage{xcolor}
\usepackage{caption}
\usepackage{subcaption}
\usepackage[%
    backend=biber,
    bibstyle=ieee,
    citestyle=ieee,
    date=year,
    isbn=false,
    doi=false,
    url=false,
    eprint=false,
    maxbibnames=99]{biblatex}
\begin{document}

%\title{Evaluation of Lidar and Visual Simultaneous Localization and Mapping (SLAM)\\
\title{Evaluation and comparison of eight popular Lidar and Visual SLAM algorithms
}

\author{\IEEEauthorblockN{1\textsuperscript{st} Bharath Garigipati}
\IEEEauthorblockA{\textit{Engineering and Natural Sciences} \\
\textit{Tampere University}\\
Tampere, Finland \\
bharath.garigipati@tuni.fi}
\and
\IEEEauthorblockN{2\textsuperscript{nd} Nataliya Strokina}
\IEEEauthorblockA{\textit{Information Technology and Communication} \\
\textit{Tampere University}\\
Tampere, Finland \\
nataliya.strokina@tuni.fi}
\and
\IEEEauthorblockN{3\textsuperscript{rd} Reza Ghabcheloo}
\IEEEauthorblockA{\textit{Engineering and Natural Sciences} \\
\textit{Tampere University}\\
Tampere, Finland \\
reza.ghabcheloo@tuni.fi}
}

\maketitle

\begin{abstract}
%Accurate localization is essential for autonomous operation of mobile robots, hence it is important to use the most suitable sensor and Simultaneous Localization and Mapping (SLAM) system based on the surroundings to navigate in an unknown environment. 
In this paper, we evaluate eight popular and open-source 3D Lidar and visual SLAM (Simultaneous Localization and Mapping) algorithms, namely LOAM, Lego LOAM, LIO SAM, HDL Graph, ORB SLAM3, Basalt VIO, and SVO2. We have devised experiments both indoor and outdoor to investigate the effect of the following items: i) effect of mounting positions of the sensors, ii) effect of terrain type and vibration, iii) effect of motion (variation in linear and angular speed). We compare their performance in terms of relative and absolute pose error. We also provide comparison on their required computational resources. We thoroughly analyse and discuss the results and identify best performing system for the environment cases with our multi-camera and multi-Lidar indoor and outdoor datasets. We hope our findings help one to choose a sensor and the corresponding SLAM algorithm combination suiting their needs, based on their target environment.
\end{abstract}

%\begin{IEEEkeywords}
%component, formatting, style, styling, insert
%\end{IEEEkeywords}

\section{Introduction}
Localization and mapping \autocite{Barfoot2017} play key roles in various applications, such as, unmanned aerial vehicles \autocite{UAV}, unmanned ground vehicles \autocite{LEGO_LOAM}, autonomous cars \autocite{kato2018autoware}, service robots \cite{service_robots}, virtual and augmented reality. These technologies are essential components of autonomous robots. They allow a robot to build a map of the environment and to track its relative position within the map. Using the map and the pose information, the robot can perform path planning, navigation, obstacle avoidance, etc. 

Nonlinear state estimation methods \autocite{Barfoot2017} such as Extended Kalman Filter (EKF) and Unscented Kalman Filter (UKF) using wheel odometry, Inertial Measurement Unit (IMU), and Global Navigation Satellite System (GNSS) are commonly used to estimate the pose of the robot \autocite{ros_loc}. However, the Kalman filter variants cannot approximate too complex nonlinear functions. Moreover, the GNSS signal is not always available, for example, indoors and among high buildings and dense forrest. Progress in Simultaneous Localization and Mapping (SLAM) \autocite{cadena2016past}  holds a promise to perform robust localization and mapping in previously unexplored environments using observations from robot's onboard sensors such as cameras and Lidars. SLAM methods are suitable for both indoors and outdoors.  %But through developments in Simultaneous Localization and Mapping (SLAM) systems, there has been a promise of increased robustness in localization in previously unexplored environments in both indoors and outdoors case, all while generating a map of the newly visited environment utilizing sensors such as cameras and Lidars. 
\begin{figure}
         \centering
         \includegraphics[width = 0.48\textwidth]{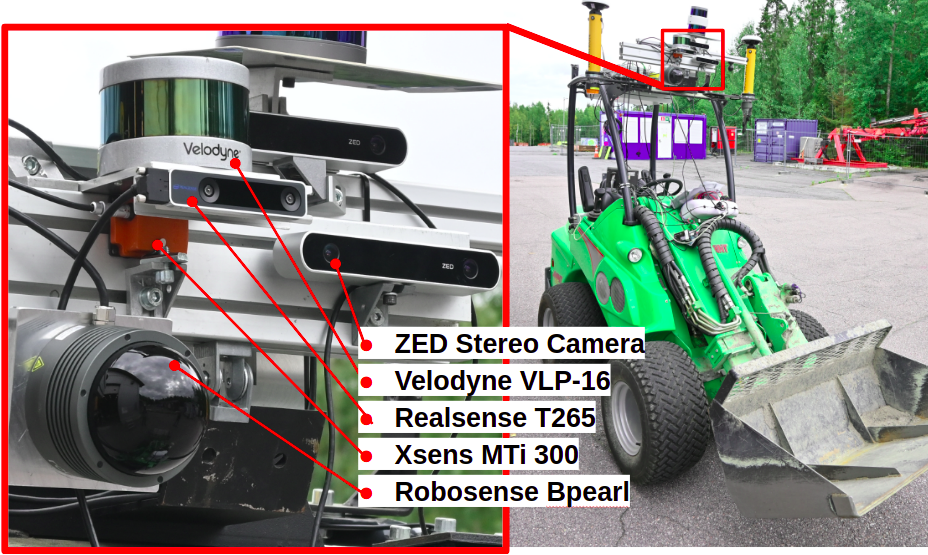}
     \caption{GIM machine used for outdoor data collection with sensor setup mounted on top.}
     \label{fig:avant1}
\end{figure}

While a number of state-of-the-art SLAM approaches exists \autocite{LEGO_LOAM, HDL_graph,8894002_ORB_SLAM3}, to the best of our knowledge there is no systematic comparison. In addition, available public benchmark datasets are either meant for testing visual-inertial SLAM only \autocite{schubert2018vidataset} or do not include different type of sensors \autocite{kitti}. However, different SLAM algorithms utilize different sensor models and types. The goal of this paper is to evaluate and compare the most common state-of-the-art SLAM algorithms using the data from different perception sensors collected simultaneously by robotic set-ups both indoor and outdoor. For a fair comparison of the performance, we used a multi-sensor setup (see Fig \ref{fig:avant1}) including two Lidars, two stereo cameras, a dual antenna GNSS (Global Navigation Satellite System), and a nine-axis inertial measurement unit (IMU) for data collection. The collected data is used to evaluate the 6 degrees of freedom (DoF) localization performance through absolute pose error (APE) and relative pose error (RPE) with respect to the reference ground truth trajectories. We analyze how the design choices made in these SLAM systems affect the accuracy of the trajectory generated by them.

%During data collection experimentation, we test multiple challenging scenarios to compare how the selected algorithms perform.  it is only fair to compare the algorithms on the data collected simultaneously on necessary sensors during experimentation. Hence, for a fair comparison of the performance, we used a multisensor setup including two Lidars, two stereo cameras, a global positioning system (GPS), and a nine-axis inertial measurement unit (IMU) for data collection. The collected data is used to evaluate the 6 degrees of freedom (DoF) localization performances through absolute pose error (APE) and relative 2 pose error (RPE) with respect to the reference ground truth trajectories, we look into how the design choices made in these SLAM systems affect the accuracy of the trajectory generated by them.
\begin{figure*}[h!]
    \centering
    \includegraphics[width=0.75\textwidth]{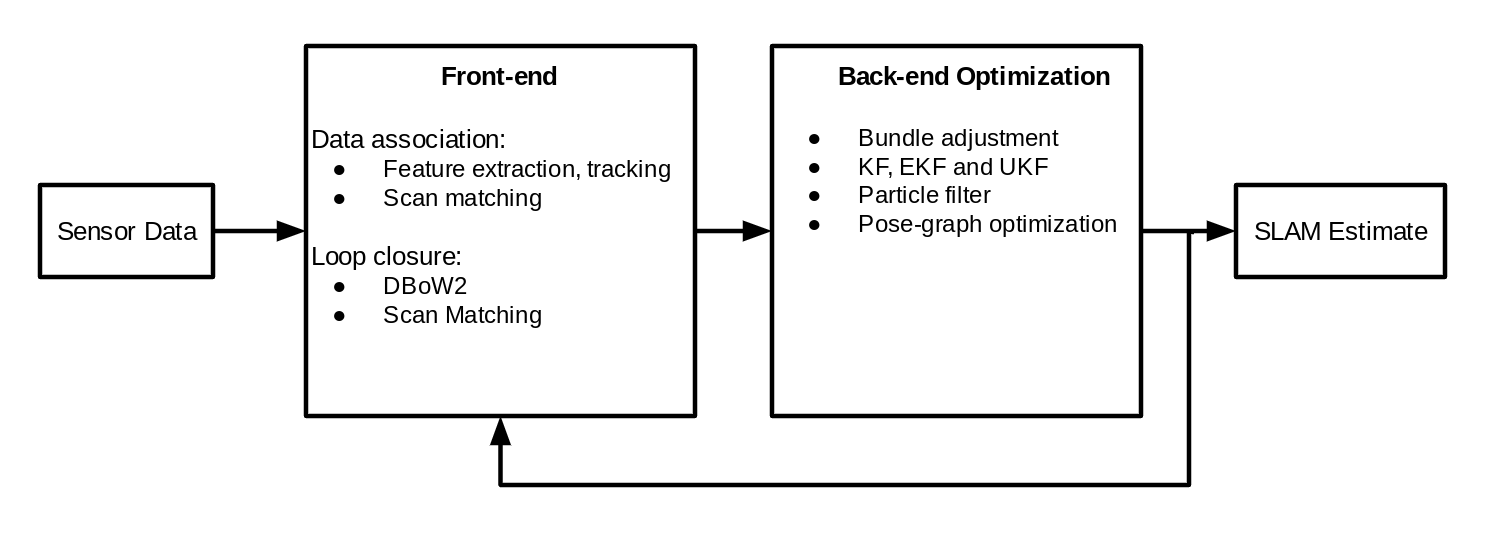}
    \caption{Components of SLAM system (based on \autocite{cadena2016past})}
    \label{fig:components_of_slam}
\end{figure*}

%\section{ Related Work}

SLAM Benchmarks suites such as Kitti \autocite{kitti} cover urban scenes including highways, rural roads, moving cars, and pedestrians. It has been an essential platform to compare the localization performance of many popular SLAM systems. When selecting a SLAM system for industrial machines, such as autonomous construction machines or other off-road machines,  several additional factors have to be considered. These factors include different  sensor mounting locations, effects of terrain type, and vibration etc. These effects are not represented in the existing benchmarks. Moreover, to the best of our knowledge, industrial settings are not covered. In this work, we build our set-up to address the effects which have not been covered in the literature. We also build a sensor suite which allows to test both Lidar and visual SLAM methods with data collected simultaneously (with different types of sensors). Some public datasets offer 3D Lidar, stereo camera, and IMU recorded simultaneously (for example, Beach Rover, ETH-challenging, Multi Vech Event \autocite{datset_list}), but they are usually recorded without a change of environment and position of sensor mount is not changed. We conduct experiments both indoor and outdoor with varying terrain type. The experiments were conducted during the research for thesis work \autocite{Bharath2021}.

Our contributions are as follows. (i) We devise indoor and outdoor experiments to systematically analyse and compare eight popular Lidar and Visual SLAM; (ii) Our experiments are devised to evaluate and compare the performance of the selected SLAM implementations against the mounting position of the sensors, terrain type, vibration effect, and variation in linear and angular speed of the sensors. (iii) We provide comparison on the required computational resources. (iv) We summarize our findings as a collection of recommendations

The paper is organized as follows. In Section II, we explain SLAM, provide some details of the internal working of the selected SLAM algorithms, and introduce the sensor suite. In Section III, we detail the experiment and the environment. In Section IV, we discuss the results. Finally in In Section V we present the conclusions. 
%\textcolor{purple}{please check the section numbers}
%It is good now
% \usepackage{graphicx}
%\begin{table}[tp]
%\resizebox{0.48\textwidth}{!}{%
%\begin{tabular}{|c|c|c|c|c|}
%\hline
%Dataset          & Affiliation & Year & Environment    & Ground truth pose \\ \hline
%Multi Vech Event & Upenn       & 2018 & Urban          & Yes               \\ \hline
%Beach Rover      & TEC-MMA     & 2017 & Terrain        & Yes               \\ \hline
%Oxford-robotcar  & Oxford      & 2016 & Urban          & Yes               \\ \hline
%KITTI            & KIT         & 2013 & Urban          & Yes               \\ \hline
%ETH-challenging  & ETH-ASL     & 2012 & Urban, Terrain &                   \\ \hline
%\end{tabular}%
%}
%\caption{Datasets with 3D LIDAR, Stereo Camera and IMU.\autocite{datset_list}}
%\label{tab:datasets}
%\end{table}

%\section{Introduction to Lidar and Visual SLAM}
%\subsection{Typical structure of SLAM}
\section{Method}

In this section, we describe main components of SLAM algorithms referring to Fig 1, we then explain the SLAM algorithms we have selected to be evaluated referring to Fig 2, and then we explain them into the detail that are relevant to our evaluation purpose.
\begin{figure*}[h]
    \centering
    \includegraphics[width=\textwidth]{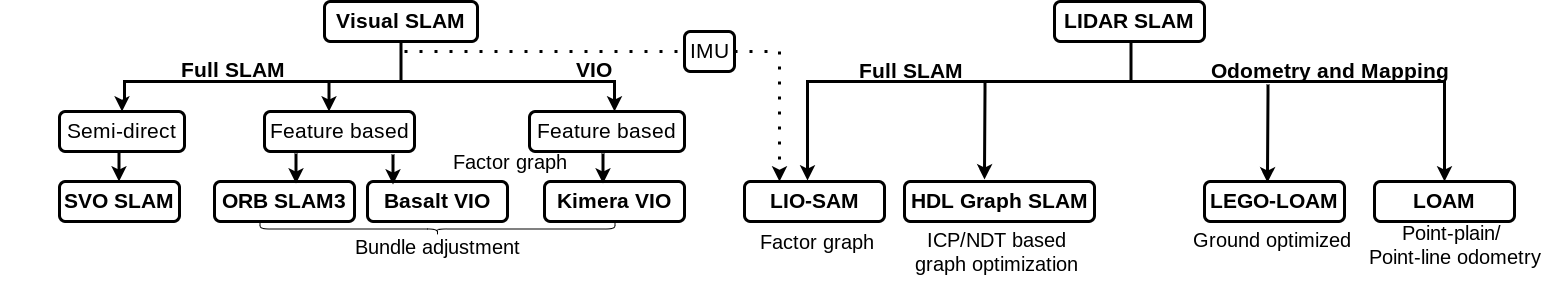}
    \caption{Selected state-of-the-art SLAM algorithms and their tested functionality.}% \textcolor{blue}{Try to make the fonts of all words inside the figure larger. Lets keep it for two columns width, but we can shorten some vertical arrows and thus make it less tall. When I read the word "selected", I wonder are those somehow specially selected? or are they most common? or stat-of-the-art? It will sound better if this is explained (for example, most common state-of-the-art). What is here tested functionality?\textcolor{red}{some of them have newer version with more capabilities, kimera also supports loopclosure through additional package, but what I tested was VIO }}}
    \label{fig:SLAM_tree}
\end{figure*}
\subsection{SLAM background}

The goal of a 3D SLAM system is to estimate the robot’s 6 DoF pose while simultaneously mapping the new environment using the input sensor data such as images and point clouds. A typical SLAM system consists of two main components: the front-end and the back-end \autocite{cadena2016past} (See Fig.~\ref{fig:components_of_slam}). 

The front-end processes the incoming visual sensor data extracting useful information from the frames which is then used for robot pose estimation. This step is usually referred to as data association. Visual SLAM front-end receives camera images, extracts key points in each frame \autocite{rublee2011orb}\autocite{rosten2008faster} and tracks them \autocite{tomasi1991detection} to match the key points between frames. An initial estimate of the robot's pose can also be generated using methods such as  \autocite{5point}. Key points consistent between the frames are called landmarks. Note that there can be additional constraints in choosing landmarks.

The front-end of Lidar SLAM typically consists of three parts: (i) the point cloud down-sampling to reduce computation, (ii) key point extractions commonly based on the smoothness value of the point cloud voxels \cite{zhang2014loam}, and (iii) scan matching such as variants of Iterative Closest Point (ICP) \cite{ICP} to generate an initial estimate of the pose transform. To reduce the localization error some systems include loop closure by visiting previously visited location. Map is typically represented as either a sparse information on landmark locations or dense point cloud representations in case of Lidar SLAM. 
%\textcolor{blue}{There are words about loop closure in the Figure. Shall we say smth about it? And what is DBoW2}

In visual SLAM, the initial pose estimate and the landmark associations from the front-end are utilized in the back-end to perform a maximum-a-posteriori estimate of the robot’s state. In case of Lidar SLAM, the algorithms use odometry from point cloud scan matching instead of landmark associations. Lidar SLAM can also use IMU integration between frames.   Recently, with the availability of libraries such as G2O \autocite{g2o} and GTSAM \autocite{-2015-GTSAM}, the most popular back-end approach is to use of factor/pose  graphs for optimization. It allows users to integrate a wide variety of sensor modalities for more robust state estimation. We present further the details on each of the selected algorithms.

%In feature-based visual SLAM methods, bundle adjustment \autocite{lourakis2009BA} is one of the most common optimization techniques.

%\textcolor{blue}{a comment about what other types there is, there was this semi-direct}
%\subsection{Properties of Perception Sensors}
%*what are the advantages and disadvantages of one sensor over the other, and they can be complimentary to each other.*
%\textcolor{red}{This section maybe not needed}

\subsection{Selection of SLAM algorithms}
We picked eight SLAM and odometry algorithms in total, to run our experiments in this paper. They are among the current state-of-the-art and widely used publicly available systems with a version of ROS-based implementations. The algorithms are split into four Lidar-based and four visual algorithms as shown in Fig \ref{fig:SLAM_tree}, within them there are both full SLAM and odometry type of algorithms. The essential difference between these is that odometry performs its estimation incrementally frame-by-frame and sometimes performs windowed local optimization, whereas full SLAM approaches aim to maintain global consistency by including loop-closure detection for detecting revisited places for correcting the error in pose estimate. This by design makes Full SLAM approaches computationally heavier than odometry algorithms.

Among the Lidar-based, there are two full SLAM approaches, including LIO SAM \autocite{9341176_LIO_SAM} and HDL graph SLAM \autocite{HDL_graph} which have loop closure correction, additionally, we used complementary IMU with LIO SAM. The two Lidar odometry algorithms that were used are  LEGO LOAM \cite{LEGO_LOAM} and original LOAM \autocite{zhang2014loam} based implementation A-LOAM \autocite{aloam}.  We used three full SLAM visual algorithms including SVO2 \autocite{Forster17troSVO}, ORB SLAM3 \autocite{8894002_ORB_SLAM3}, Basalt VIO \autocite{usenko2019basalt}, and odometry implementation of Kimera VIO \autocite{Kimera}.

\subsection{Algorithms Overview}
\noindent\emph{Lidar SLAM:} In Lidar based algorithms we included LOAM as it is the building block for many popular Lidar-based SLAM systems including LIO SAM and LEGO LOAM. It is a Lidar odometry algorithm that uses point-to-line and point-to-plane variants of ICP to perform scan match odometry. LEGO LOAM is the other Lidar odometry that we tested, it is very similar to LOAM but achieves efficiency gain by splitting the point cloud into edge and plane features. The algorithm separately extracts height, roll, and pitch from planar features (ground), and X-, Y-coordinates and yaw from edge features.  

One of the Lidar-based full SLAM algorithms tested was HDL graph SLAM. It is a factor graph-based approach which allows the user to define multiple edge constraints such as GPS, IMU, floor plane detection, and loop closure. Nodes are added to the graph structure after initial scan matching which is chosen by the user to be either ICP or NDT method. LIO SAM is the second full SLAM Lidar technique we tested. It estimates states utilizing tightly coupled IMU integration and Lidar odometry via factor graph optimization.

\noindent\emph{Visual SLAM:} First in the Vision-based SLAM system is SVO2. Unlike other Visual SLAM algorithms which use keypoint detectors on the front-end, SVO2 has a front-end similar to direct visual SLAM system. There, the estimate is based on sparse model-based image alignment using the local intensity gradient’s direction to estimate the motion. This gives it a faster front-end compared to feature-based systems. Among feature-based SLAM we used ORB SLAM3 which is considered to be one of the current state-of-the-art visual SLAM systems. Its front-end consists of the ORB feature detector and descriptor matching and it offers extra robustness under demanding conditions with its multimap data association DBoW2 posegraph-based loopclosure.

Basalt VIO is the next feature-based visual-inertial algorithm that we used. Basalt recovers the non-linearity in the IMU reading between keyframes which are lost during the more general IMU pre-integration step. The local bundle adjustment estimate generated using FAST features and KLT tracker along with non-linear factors are optimized during global bundle adjustment using factor graphs. Finally, Kimera Visual-inertial odometry from the Kimera C++ library, was developed to perform semantic 3D mesh reconstruction of the environment using metric semantic SLAM. its front-end begins with Shi-Tomasi corners \autocite{tomasi1991detection} tracked through the new image frames using the KLT tracker. A five-point/three-point RANSAC is then used to solve the relative pose between keyframes, and pre-integrated IMU measurements are calculated between keyframes. The landmarks, initial pose estimate, and IMU pre-integration are used as factor graph constraints to generate the final state estimate.

\section{Experiments description}
In this section, we describe our set-up and the experiments we have conducted to study the effect and severity of following items on accuracy and reliability of the selected SLAM implementations: 
\begin{itemize}
    \item[i] mounting position of the sensors,
    \item[ii] terrain type and vibration effect, and 
    \item[iii] variation in linear and angular speed of the sensors. 
\end{itemize}
In outdoor scenarios, we study the effects of all three items, and for indoor case our experiments focuses on (iii). %\textcolor{purple}{It will also be good to have some references to refer where the other authors has mentioned that such things have negative/positive affect.}
% \textcolor{orange}{To do: list sources where the effects were mentioned}

%\subsection{Sensor suite}
%\begin{figure}[h!]
%    \centering
%    \includegraphics[width=0.43\textwidth]{figures/sensor_setup.png}
%    \caption{Sensor setup}
%    \label{fig:sensor_bar}
%\end{figure}

%\vspace{\baselineskip}

\subsection{Experiment set-up}

\begin{figure}
         \centering
         \includegraphics[width = 0.35\textwidth]{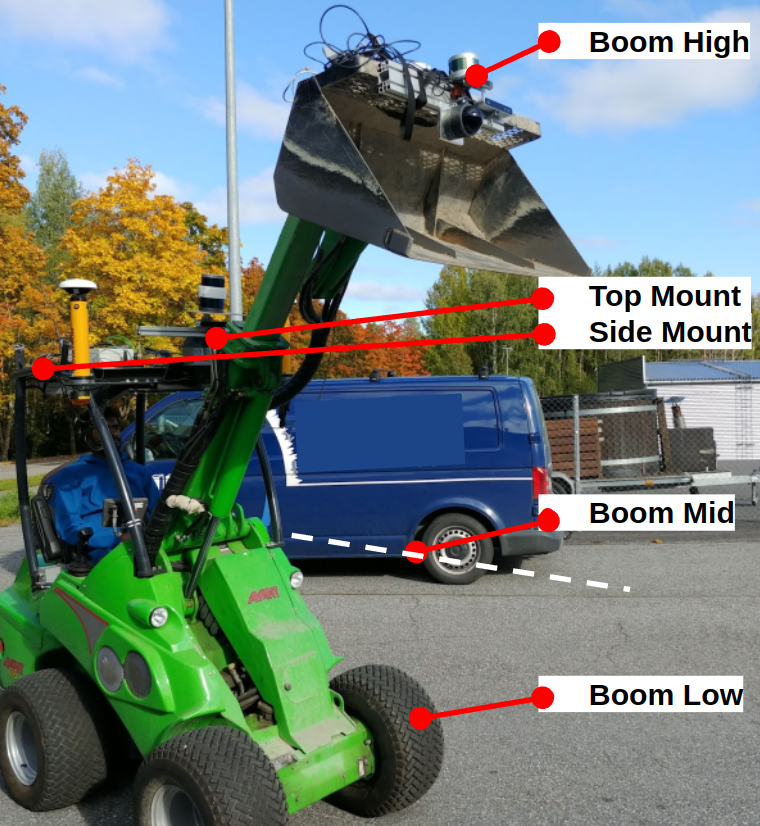}
         \caption{Sensor mounting positions for outdoor experiments.}
         \label{fig:avant}
\end{figure}

\begin{figure}

         \centering
         \includegraphics[width=0.4\textwidth]{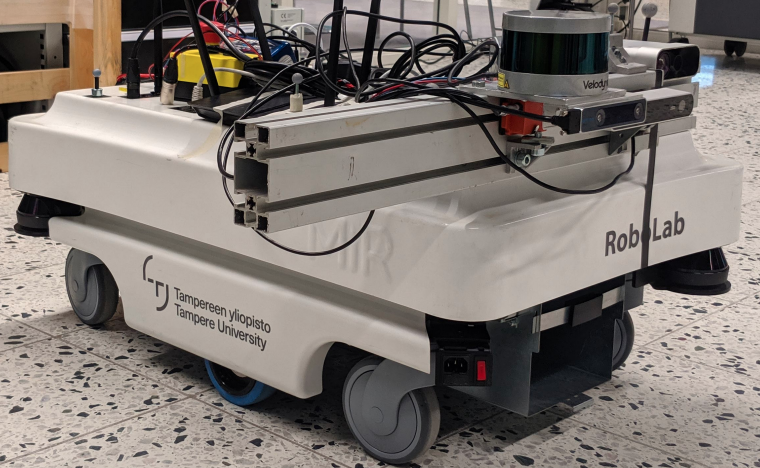}
         \caption{MiR robot used in indoor experiments.}
         \label{fig:mir}
\end{figure}

\noindent\emph{Sensor suite and computation:}

In this paper two Lidars, two stereo cameras, and an IMU are used in the sensor setup in Fig \ref{fig:avant1}. Velodyne VLP-16 is a rotating 16 scan Lidar with a vertical field of view of 30 degrees; it offers a vertical resolution of 2 degrees and a horizontal resolution of 0.1 to 0.4 degrees dependent on a rotating speed ranging from 5 to 20 Hz. The second Lidar is the robosense Bpearl; it is likewise a spinning Lidar with a higher 32 scan lines, it has a broader vertical field of view of 90 degrees. With a resolution of 2.81 degrees and a horizontal resolution of 0.2 to 0.4 degrees depending on the speed, which ranges from 10 to 20 Hz. Both Lidars have a maximum range of 100m and a precision of ±3cm. Among the two stereo cameras, Intel Realsense T265 is a global shutter monochrome fisheye camera with a built-in IMU, it has a baseline of 64mm and a field of view of 163 degrees. The second stereo camera is a rolling shutter RGB camera with a larger baseline of 120mm and a vertical and horizontal field of view of 60 and 90 degrees respectively. Finally, the IMU utilized was 9-axis Xsens Mti 300. During the experiments, sensor data was recorded into ROS bag files on a Jetson Xavier NX  module with an m.2 SSD running ROS melodic.

In our experiments, Velodyne VLP 16 was used with A-LOAM, LEGO LOAM, HDL graph SLAM, and LIO SAM. We used Realsense T265 with Basalt VIO, SVO2, and Kimera VIO. Kimera VIO did not support the ultra-wide-angle fish-eye model of T265 perfectly and failed in outdoor experiments. Finally, we used the ZED camera with ORB SLAM3. As a result of initial testing, we found these combinations of sensors and algorithms to have the best possible results. 
We used the same sensor suite in both indoor and outdoor experiments.

The algorithms were running on a 4 core 8 thread Intel i7 7700HQ CPU. In our evaluation, the max capacity of all 8 threads together is considered as 100\%. During run-time, the overall usage data is saved into a text file, which is later parsed to separate and calculate the usage corresponding to processes started by the Falgorithms. 
%\textcolor{purple}{All the computations are performed on (somePC?) and the something about ROS implementations}
\vspace{\baselineskip}
\noindent\emph{Outdoor environment:}
The outdoor tests were conducted at the Mobile Hydraulics Lab of Tampere University. As a ground truth, we used our custom built localization based on dual antenna RTK-GNSS, IMU, and odometry. The sensors were assembled on a frame, and  calibrated using Kalibr toolbox \autocite{rehder2016extending} for camera-imu calibration and LI-Calib \autocite{lv2020targetless} for Lidar-imu calibration . The sensor frame was mounted on top of GIM machine (a multi-purpose wheel loader modified in Tampere University for the purpose of research in autonomous operations- ref MED 09), see Fig \ref{fig:avant}.
%[ref- we can refer to our IROS paper as well]
\vspace{\baselineskip}
\noindent\emph{Indoor environment:} The indoor tests were conducted in RoboLab Tampere of Tampere University. To obtain consistent predefined motion in these trajectories, we mounted the sensor bar on Mobile Industrial Robot (MIR) shown in Fig \ref{fig:mir} it is a commercial differential drive mobile robot running Robot Operating System (ROS), generally used to move material within an industrial warehouse space. Through its interface, MIR allows users to define a path that it will execute; this is useful to design and perform repetitive motions for testing SLAM algorithms. Tampere university’s Robolab is equipped with the state-of-the-art Optitrack motion capture system used to collect ground truth. It was used to collect the ground truth trajectory by tracking the markers placed on MIR. 
\vspace{\baselineskip}
%\noindent\emph{SLAM implementations:} We have used original implementations of the mentioned algorithms, with minor modifications in some cases. The sources and some critical hyper-parameters are summarised in table XX.

%w move to the experimentation stage where we test a few challenging cases for the SLAM system in indoor and outdoor using the data collection setup introduced in the previous section. We first go through the conducted outdoor experiments using the GIM machine where have changing terrains and speeds and compare the localization performance of the SLAM algorithms, then we look at the indoor experiments conducted with the MiR robot and their localization results. Finally, we compare the CPU usage of the chosen algorithms to evaluate their efficiency. 
%\textcolor{blue}{Can we already here in few sentences introduce the specifics of the outdoor and indoor tasks and why they were selected. What challenge they contain}
%\nataliya: Reza now addressed it
%\textcolor{blue}{Somewhere should be said that where the computations were happening}
%\nataliya: space for this was also created

%\textcolor{blue}{as Reza commented, shall we merge Fig 3 and 4};

\begin{figure}[h]
    \centering
    \includegraphics[width=0.47\textwidth]{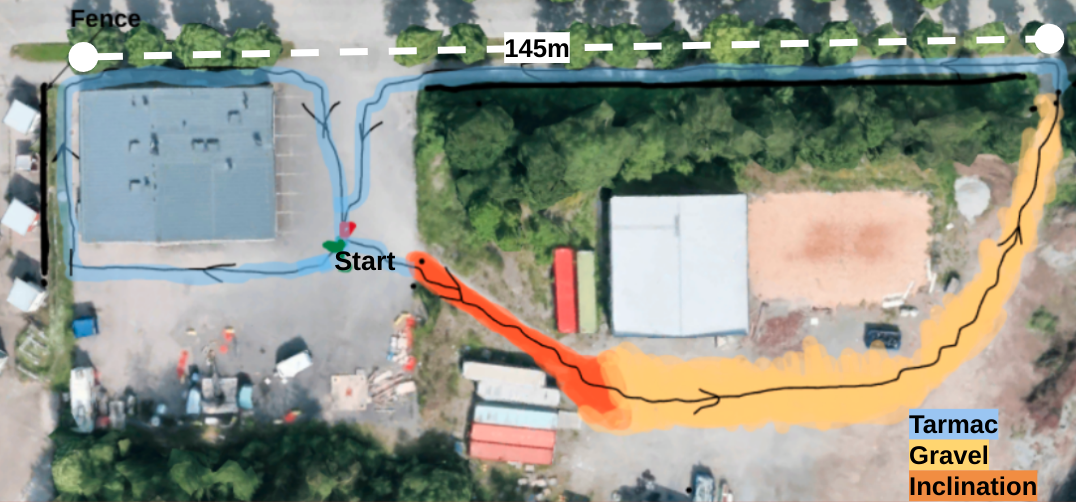}
    \caption{Outdoor experiment 1: top of view of the trajectory to test the effect of various mounting positions and terrain types. }
    \label{fig:terrain_change}
\end{figure}
\begin{figure}[h]
    \centering
    \includegraphics[width=0.45\textwidth]{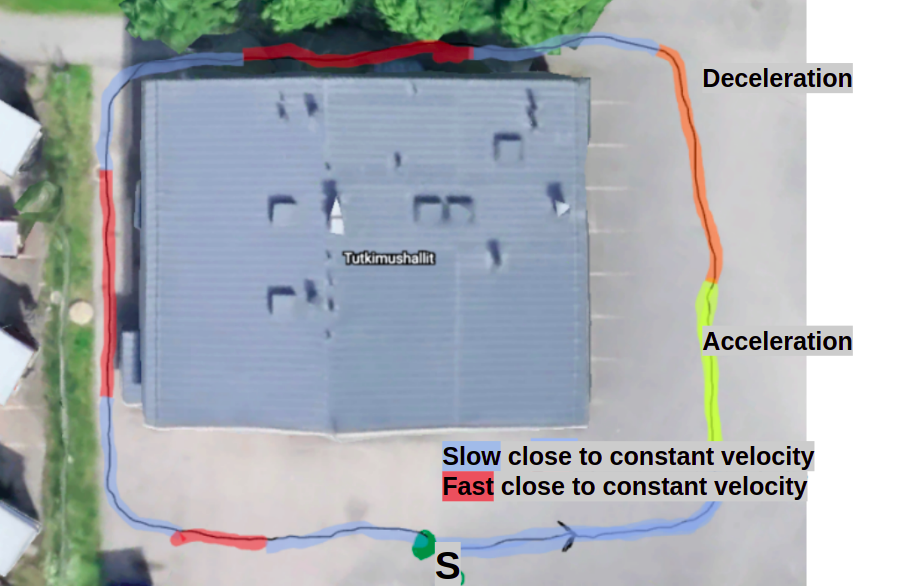}
    \caption{Outdoor experiment 2: top view of of the trajectory to test the effect of variation in linear and angular speed. }
    \label{fig:speed_change}
\end{figure}

%\textcolor{blue}{Could you enlarge the text in this figure. DONE}
%\begin{figure}[h]
%    \centering
%   \includegraphics[width=0.4\textwidth]{figures/mir.png}
%    \caption{MiR robot with sensor setup attached. \textcolor{blue}{the same here, could we mark the sensors and possibly their names.}}
%    \label{fig:mir}
%\end{figure}
\begin{figure}[bp]
    \centering
    \includegraphics[width=0.47\textwidth]{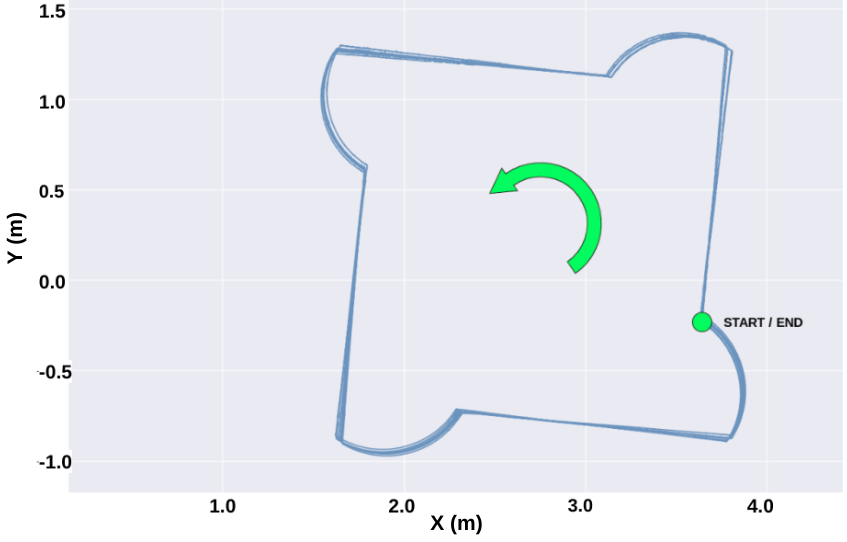}
    \caption{Indoor experiment 1: ground truth sensor trajectory. Notice that the sensor is mounted on the right corner of the MiR robot}
    \label{fig:square_loop}
\end{figure}

\begin{figure}[tp]
    \centering
    \includegraphics[width=0.47\textwidth]{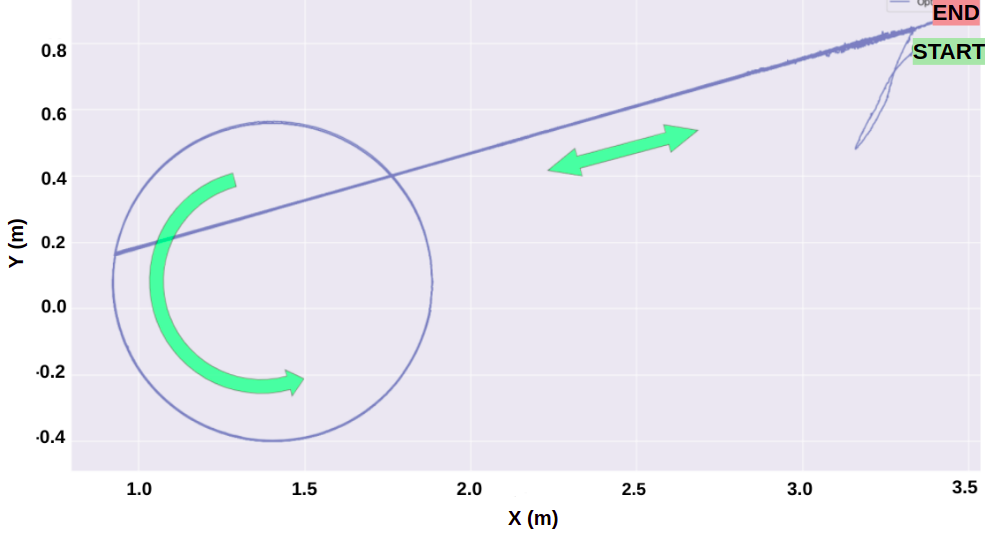}
    \caption{Indoor experiment 2: Optitrack ground truth sensor trajectory.}
    \label{fig:360_pan}
\end{figure}

\subsection{Outdoor Experiments}
%\textit{Set-up --} The outdoor tests were conducted at the mobile hydraulics lab of Tampere University. The sensor setup, consisting of two Lidars, two stereo cameras, and IMU as described earlier, was mounted on GIM machine shown in Fig~\ref{fig:avant} which is a modified Avant 600 multipurpose loader with various sensors and controllers installed. The IMU, GNSS, and odometry-based ekf localization from GIM machine is collected as a reference trajectory. 
Two outdoor experiments were designed to study the effects i-iii mentioned above: one set to study the effect of mounting positions and terrain types, and the second set to study the effect of speed of motion.

\noindent\emph{Various mounting positions and terrain types (Outdoor experiment 1):} 
%\textcolor{blue}{I wonder if we shall put the terrain figures into one textwide figure as subfigures. It somehow looks better I think when they are side by side}. 
These experiments contain five datasets where the machine repeats the same trajectory in the shape of an eight loop around Mobile Hydraulics LAB as shown in Fig ~\ref{fig:terrain_change}. In each test, the sensor suite is mounted in different position of the machine (5 positions in total shown in Fig ~\ref{fig:avant}): 1) on the boom-low, 2) on the boom-middle, 3) on the boom-high, 4) top mount on the cabin frame, 5) side mount. 
 This experiment allows us to test how the changes in terrain affect the localization performance and benefits of having loop closure after passing through a challenging environment. We also test the sensitivity to vibration effects and sensor mounting position.
The trajectory includes straight line on asphalt, straight line on gravel, and downhill path on gravel terrain.

%\textcolor{blue}{would be really good to mark them in the avant figure. Is it important to know the approximate height in m?}  \textcolor{red}{they weren't measured, so it would just be a rough estimate, is that still ok? check image sensor postions}.
%\textit{Terrain type and mounting position--} this experiment contains five datasets where  the Avant loader repeats the same trajectory in the shape of an eight loop around Mobile Hydraulics LAB as shown in Fig ~\ref{fig:terrain_change}. The trajectory includes straight asphalt, straight gravel, and downhill gravel terrain types. 
%\textcolor{blue}{I wonder if we shall put the terrain figures into one textwide figure as subfigures. It somehow looks better I think when they are side by side}. 

%having the same dataset with different mounting positions should also help us find the effects of mounting positions on the accuracy of localization.

%\begin{figure}[bp]
%    \centering
%    \includegraphics[width=0.45\textwidth]{figures/avant.png}
%    \caption{GIM machine used for outdoor data collection with sensor setup mounted on top. \textcolor{blue}{Let's try to get a more wide but less tall horizontal picture and mark on it the sensors. See, for example figures/setup.jpg - this is from my paper and maybe quality not so good.}}
%    \label{fig:avant}
%\end{figure}

\vspace{\baselineskip}

\noindent\emph{Variation in linear and angular speed (Outdoor experiment 2):} In this experiment, we test the rapid changes of machine speed on the localization accuracy of the SLAM algorithms. Sensor motion introduces skew distortion in accumulated point clouds from a Lidar, and motion blur in images where slower feature detectors in the visual SLAM front-end can lose track of the landmarks. This dataset was collected at the same location as the shorter leg of the previous terrain change dataset. The GIM machine starts at the start location S, and it goes around counterclockwise as shown in the Fig~\ref{fig:speed_change}. During this single loop, the machine goes through changes in velocity such as constant velocity, acceleration, and braking; the loop ends at where it started. 

\subsection{Indoor Experiments}
%\textit{Set-up --} The indoor experiments were recorded at Robolab of Tampere University, using the same sensor setup as in outdoor experiments. To obtain consistent predefined motion in these trajectories, we mounted the sensor bar on Mobile Industrial Robot (MIR) shown in Fig \ref{fig:mir}. MIR is a commercial differential drive mobile robot running Robot Operating System (ROS), generally used to move material within an industrial warehouse space. Through its interface, MIR allows users to define a path that it will execute; this is useful to design and perform repetitive motions for testing SLAM algorithms. Tampere university’s Robolab is equipped with the state-of-the-art Optitrack motion capture system described in section (ref optitrack). It was used to collect the ground truth trajectory by tracking the markers placed on MIR.

We designed three experiments to study how robust the low-level tracking components are and potential drift in various situations, namely, looping in square, straight line with 360 deg turn, stationary with dynamic scene. 
\vspace{\baselineskip}

\noindent\emph{Looping in square (Indoor experiment 1):} In the first indoor experiment, MiR was assigned to follow a square-shaped path with two-meter sides in a continuous loop for five minutes as shown in Fig \ref{fig:square_loop}, total of ~72 meters. This dataset is intended to observe the possible drift in the localization in a constant loop that has panning right-angle turns at each corner of the square, which might have a potential effect on the ability of the feature tracker. Two of the four walls in the Robolab test area are made of glass, with a view of the wider lobby area of the building; this can result in partial reflection of Lidar beams and occlusion due to the objects in front of the glass possibly affecting Lidar SLAM performance.
\begin{table*}[h!]
\centering
\resizebox{0.7\textwidth}{!}{%
\begin{tabular}{|c|c|c|c|c|c|}
\hline
{Mounting position} & Boom Low       & Boom Mid       & Top Mount      & Side Mount     & Boom High \\ \hline
LOAM       & 2.062          & 3.208          & 1.751          & \textbf{1.669} & 5.918     \\ \hline
LEGO LOAM  & 1.316          & 2.759          & 11.853         & 8.647          & 7.891     \\ \hline
LIO SAM    & 2.157          & 10.778         & \textbf{1.142} & 1.194          & 3.467     \\ \hline
HDL Graph  & 1.99           & \textbf{1.474} & 1.847          & 3.983          & 3.337     \\ \hline
ORB SLAM3  & \textbf{1.563} & 3.014          & 8.641          & 29.922         & 13.046    \\ \hline
Basalt VIO & \textbf{2.679} & 3.360          & cam fail       & failed         & 7.094     \\ \hline
SVO2       & \textbf{4.559} & 4.662          & cam fail       & failed         & 5.589     \\ \hline
\end{tabular}%
}
\caption{Outdoor experiment 1: RMS of Absolute Pose Error (APE) in meters for different mounting positions}

%\textcolor{purple}{what the boom high is not after Boom mid?} \textcolor{brown}{boom high was higher than top mount and side mount, this order is to show increasing height from left to right (for example in ORB, you can see the increasing error from left to right)}

%\textcolor{blue}{This concerns all the tables 1-4: (i) either we need to make them span over two columns (for example like this) or to transpose them so that narrower dimension would be horizontal and longer vertical, otherwise too small letters; (ii) is it possible to correct that the Table title goes a bit on the table; (iii) it is also a good practice to have same number of number after comma in all numbers and to think how many actually are needed to capture the difference (but this is a small thing). }
\label{tab:terrain_change}
\end{table*}

\vspace{\baselineskip}

\noindent\emph{Straight line with a 360 turn (Indoor experiment 2):} In the second indoor experiment, MiR moves forward on a straight line, starts rotating counterclockwise until it reaches 360 degrees, and returns to the start of the rotation. After this, it returns back in a straight line to the starting point, and repeats the motion with clockwise rotation in the alternative turns, the top view of the trajectory can be seen in Fig \ref{fig:360_pan}. The experiment is recorded for five minutes. The 360-degree rotation is meant to induce skew in the point cloud and see how different Lidar SLAMs deal with it. The features are also closer to the sensors as it is indoors resulting in fast moving key-points, rolling shutter distortion and motion blur during the rotation, which could be challenging to slower feature trackers in the visual SLAM algorithms. Additionally, the algorithms would have to re-localize accurately in case of lost features.

\vspace{\baselineskip}

\noindent\emph{Dynamic scene (Indoor experiment 3):} The MiR robot is kept stationary throughout the third experiment, but there are moving objects in front of the sensors such as pallets, chairs, and people. The experiment was recorded for five minutes. This dataset allows to measure how dynamic objects affect the accuracy of the SLAM estimate by measuring the drift in the generated pose.

%\textcolor{blue}{Could you verify that this trajectory goes like that? Did it have circle only on one side? Could you enlarge the words START and END}
\section{Results}
\noindent\emph{Performance metrics:} to evaluate the performance, we compute an error between the trajectories generated by the SLAM algorithms and the ground truth. We use Root Mean Square (RMS) of Absolute Pose Error (APE) and Relative Pose Error (RPE), as well as their Standard Deviation (STD). %\textcolor{purple}{please define these terms more rigorously. RMS is clear. APE: is it just Euclidean norm and angles described by Euler, absolute part is clear it is with respect to ground truth? RPE: relative to what, to itself with or without loop-closure? if these are correct your first sentence of the paragraph is not correct.}

\begin{equation}
    APE_i = Q_i^{-1}P_i
    \label{eq:ape}
\end{equation}

\begin{equation}
    RPE_{i+\Delta} = (Q_i^{-1}Q_{i+\Delta})^{-1}(P_i^{-1}P_{i+\Delta})
    \label{eq:rpe}
\end{equation}

$Q_i$ is the ground truth transformation between ground truth origin and robot position at time $i$. $P_i$ is the estimate of robot in arbitrary SLAM origin. To obtain meaningful APE, we need to align SLAM origin and ground truth's origin, in this paper, we used least square based Umeyama alignment algorithm \autocite{Umeyama}. RPE bypasses the alignment problem by comparing the relative poses of the robot between time $i$ and $i+\Delta$  of ground truth and pose estimate.

%\textcolor{blue}{when we have also theory written with formular, I think, it would be possible and good to specify errors in formulas}
%\textcolor{blue}{You can say: first we demonstrate results for...experiments, then ...}

% \usepackage{booktabs}
% 

\vspace{\baselineskip}

\noindent\emph{Outdoor experiments:} Table \ref{tab:terrain_change} shows the RMS of APE  of the trajectories generated by the SLAM algorithms compared to the ground truth for different sensor mounting positions (Experiment 1 shown in Fig \ref{fig:terrain_change}). 

The ground-optimized  LEGO LOAM’s accuracy degrades as the elevation of the sensors increases (i.e., Boom High, Top Mount). LIO SAM, which uses IMU demonstrates the best APE of 1.142m, but its performance degrades when sensors are mounted on the shaky boom. LOAM, which is based on variants of ICP scan matching has the best accuracy in side mount position which provides the largest unobstructed field-of-view of around 270 degrees. 

 ORB SLAM3 had the least error of 1.563m among all of the visual SLAM algorithms in boom low position. It can also be seen that the accuracy of the overall visual SLAM algorithms goes down as the camera moves away from the ground. In the side mount position, as the camera is perpendicular to the motion of the machine, the key points are moving quickly and the motion blur was high. Data association between frames was rarely possible, resulting in the failure of Basalt VIO and SVO2, but OBR SLAM3 managed to relocalize due to its better data association capabilities although with large error. Realsense T265 hardware failed during the collection of the Top mount dataset, resulting in missing estimates from Basalt VIO and SVO2.

% \usepackage{graphicx}
\begin{table}[h]
\centering
\resizebox{0.5\textwidth}{!}{%
\begin{tabular}{|c|c|c|c|c|c|c|c|}
\hline
APE   & LOAM        & LEGO & LIO  & HDL  & ORB & Basalt & SVO2   \\ 
      &             & LOAM & SAM & Graph & SLAM3 & VIO & \\ \hline
RMS     & 1.072     & 1.448 & 3.715 & \textbf{0.845}& 1.442 & 1.245 & 2.919 \\ \hline
STD   & 0.480     & 0.880 & 1.701 & \textbf{0.387}& 0.487 & 0.452 & 1.433\\ \hline
\end{tabular}
}

\caption{Outdoor experiment 2: RMS and STD of Absolute Pose Error (APE)  in meters for various speed experiments.}
\label{tab:speed_change}
\end{table}

%\begin{table}[h]
%\centering
%\resizebox{0.3\textwidth}{!}{%
%\begin{tabular}{|c|c|c|}
%\hline
%APE error  & RMS        & STD   \\ \hline
%LOAM       & 1.072          & 0.480 \\ \hline
%%LEGO LOAM  & 1.448          & 0.88  \\ \hline
%LIO SAM    & 3.715          & 1.701 \\ \hline
%HDL Graph  & \textbf{0.845} & 0.387 \\ \hline
%ORB SLAM3  & 1.442          & 0.487 \\ \hline
%Basalt VIO & 1.245          & 0.452 \\ \hline
%SVO2       & 2.919          & 1.433 \\ \hline
%\end{tabular}
%}%

%\caption{RMS and STD of Absolute Pose Error for various speed experiments in outdoor}
%\label{tab:speed_change}
%\end{table}
In the second experiment (Fig \ref{fig:speed_change}) for varying linear and angular speed, HDL graph SLAM has the least error of 0.845m as shown in Table \ref{tab:speed_change}, Its loop-closure capability helped in maintaining estimate’s accuracy. The ICP scan match odometry algorithm LOAM performed well with an error of 1.072m, LEGO LOAM has a decent RMS APE of 1.448m as the boom was in boom-mid position with a good view of the ground. LIO SAM performed the worst with an error of 3.715m due to an initialization error. ORB SLAM3 and Basalt VIO had similar errors, both better than the Lidar SLAM algorithms LIO SAM and LeGO LOAM. 

%\textcolor{blue}{Can we make some general conclusion for outdoor experiments? For example, if you have a machine like Avant and outdoor what would you choose as the first option? and why? How would you locate sensors? Are those demonstrated error suitable for usage in field robotics?} \textcolor{purple}{Nataliya's comment is important. we can either have a short paragraph with heading: "Discussion" or "conclusion" to discuss these rule of thumb. or have a dedicated subsection. I like the former better.}

\begin{table*}[pht]
\centering
\resizebox{0.9\textwidth}{!}{%
\begin{tabular}{|c|ccc||ccc|| c|c|}
\hline
           & \multicolumn{3}{c|}{Square loop (Exp. 1)}                                                           & \multicolumn{3}{c|}{360 pan (Exp. 2)}             & \multicolumn{2}{c|}{Dynamic scene (Exp. 3)}                                       \\ \hline
Error       & \multicolumn{1}{c|}{RPE}        & \multicolumn{1}{c|}{RPE STD}        & APE        & \multicolumn{1}{c|}{RPE}         & \multicolumn{1}{c|}{RPE STD}         & APE & Drift from origin & Accumulated distance \\ \hline
LOAM       & \multicolumn{1}{c|}{0.030}          & \multicolumn{1}{c|}{0.011}          & 1.505          & \multicolumn{1}{c|}{0.2772}          & \multicolumn{1}{c|}{0.0129}          & 0.8316  & 0.0484 & 2.4220 \\ \hline
LEGO  LOAM & \multicolumn{1}{c|}{0.105}          & \multicolumn{1}{c|}{0.108}          & 1.507          & \multicolumn{1}{c|}{0.0972}          & \multicolumn{1}{c|}{0.0397}          & 0.8279  & 0.0058 & 3.1880\\ \hline
LIO SAM    & \multicolumn{1}{c|}{0.0606}         & \multicolumn{1}{c|}{0.037}          & 1.509          & \multicolumn{1}{c|}{0.0551}          & \multicolumn{1}{c|}{0.0263}          & 0.8293  &0.0165 & 2.2400 \\ \hline
HDL Graph  & \multicolumn{1}{c|}{0.241}          & \multicolumn{1}{c|}{0.174}          & 1.449          & \multicolumn{1}{c|}{0.3270}          & \multicolumn{1}{c|}{0.2148}          & 0.8392  & - & - \\ \hline
ORB SLAM3  & \multicolumn{1}{c|}{0.016}          & \multicolumn{1}{c|}{0.008}          & \textbf{1.388} & \multicolumn{1}{c|}{0.0232}          & \multicolumn{1}{c|}{0.0175}          & 0.8313  & 0.2345 & 0.8480\\ \hline
Basalt VIO & \multicolumn{1}{c|}{\textbf{0.008}} & \multicolumn{1}{c|}{\textbf{0.003}} & 1.412          & \multicolumn{1}{c|}{\textbf{0.0084}} & \multicolumn{1}{c|}{\textbf{0.0037}} & \textbf{0.8186}  & 0.5398 & 8.7310\\ \hline
Kimera     & \multicolumn{1}{c|}{0.029}          & \multicolumn{1}{c|}{0.012}          & 1.716          & \multicolumn{1}{c|}{0.0342}          & \multicolumn{1}{c|}{0.0214}          & 0.8811  & \textbf{0.0016} & \textbf{0.0340} \\ \hline
SVO2       & \multicolumn{1}{c|}{0.009}          & \multicolumn{1}{c|}{\textbf{0.003}} & 1.475          & \multicolumn{1}{c|}{0.0086}          & \multicolumn{1}{c|}{\textbf{0.0037}} & 0.8325  & 0.3648 & 0.4000\\ \hline
\end{tabular}%
}

\caption{Indoor experiments: RMS, STD of Relative Pose Error (RPE) and APE  with respect
to the ground truth trajectories. Values are in meters.}
\label{tab:sq_loop_and_360}
\end{table*}

\vspace{\baselineskip}

\noindent\emph{Indoor experiments:} Results of the first indoor experiment (looping in square) are summarized in Table~\ref{tab:sq_loop_and_360}.  LOAM had the smallest Relative Pose Error (RPE) of 0.030m among the Lidar based algorithms, followed by LIO SAM with an error of 0.0606m among Lidar SLAM algorithms. Visual SLAM algorithms had better RPE than Lidar SLAM, with Basalt having the least error of 0.008m. By comparing the standard deviation, it can be seen that on average Lidar SLAM estimates were noisier compared to visual SLAM estimate
The global consistency of the estimates can be interpreted through APE. ORB SLAM3 has the best APE indicating superior loop-closure and global optimization capabilities. Similarly, HDL graph SLAM had the worst RPE results but comparable APE results. It is interesting to note that three Lidar-based algorithms, LOAM, LEGO LOAM, and LIO SAM, have a very close APE of 1.5m. SVO performed better in this dataset than the previous outdoor datasets due to its faster front-end, helping it track features better through the panning 90 degree turns of the square.

%\begin{table}[tp]
%\centering
%\resizebox{0.48\textwidth}{!}{%
%\begin{tabular}{|c|c|c|}
%\hline
%Error      & Drift from origin & Accumulated distance \\ \hline
%LOAM       & 0.0484            & 2.4220               \\ \hline
%LEGO LOAM  & 0.0058            & 3.1880               \\ \hline
%LIO SAM    & 0.0165            & 2.2400               \\ \hline
%HDL Graph  & 0.0000            & 0.0000               \\ \hline
%ORB SLAM3  & 0.2345            & 0.8480               \\ \hline
%Basalt VIO & 0.5398            & 8.7310               \\ \hline
%Kimera VIO & 0.0016            & 0.0340               \\ \hline
%SVO2       & 0.3648            & 0.4000               \\ \hline
%\end{tabular}%
%}
%\caption{Drift from the starting position and accumulated distance caused by the moving objects.}
%\label{tab:indoor_stationary}
%\end{table}

% \usepackage{graphicx}

%\textcolor{purple}{I am not sure if the picture Fig 5 and specially 6is clear: text small, units missing, what is the circle, the robot, its trajectory? why there is no circle at the other end? why the line does not end at the middle of circle or or somehow symmetrically?}
%\textcolor{brown}{figure 8? senors are mounted on the right side on the front of MIR causing the offset when making left circle around MiR's Z axis. That figure is the trajectory in sensor frame. }

Table \ref{tab:sq_loop_and_360} shows the RPE and APE from the second indoor experiment in the middle, which has trajectory shown in Fig \ref{fig:360_pan}. Visual SLAM algorithms performed better than Lidar SLAM again in RPE, where Basalt has the least RMSE of 0.0084m followed by SVO2 with an error of 0.0086m. ORB SLAM3 has the slower front-end with the ORB detector, compared to the best performing Basalt which uses a less computationally costly FAST detector, ORB SLAM3 constantly lost track of features during the rotation but was able to re-localize to achieve comparable accuracy with its superior loop closure detection and submap merging.
LIO-SAM using the IMU pre-integration and IMU point cloud deskewing has the least error among Lidar SLAM with an RPE of 0.0551m. 

In the final indoor experiment, the robot is stationary with objects moving in the view, therefore we use the accumulated distance and drift from the starting position as metrics to compare the quality of the SLAM estimate as shown in Table \ref{tab:sq_loop_and_360}. The drift from starting point in Lidar-based algorithms is minimal, and they outperformed all vision-based algorithms except Kimera. Pose graph nodes have not been initialized in HDL graph SLAM as the minimum distance necessary has not been exceeded, keeping the estimate at zero. Visual algorithms ORB SLAM3, Basalt VIO, and SVO2 have drifted from the start point around 0.23-0.36m, higher than Lidar SLAM, indicating that smaller field-of-view of the camera compared to Lidar led to greater impact of moving objects. Lidar SLAM estimates were accurate but noisy resulting in the higher accumulated distance although the overall drift was lower. 

\begin{table}[hpt]
\resizebox{0.48\textwidth}{!}{%
\begin{tabular}{|c|c|c|}
\hline
Usage     & Average CPU usage (\%) & Peak CPU usage (\%) \\ \hline
LOAM      & 6.637                  & 11.361              \\ \hline
LEGO LOAM & \textbf{4.433}                  & \textbf{7.880}     \\ \hline
LIO SAM   & 21.401                 & 79.650              \\ \hline
HDL Graph & 11.85                  & 56.120              \\ \hline
ORB3      & 27.743                 & 43.319              \\ \hline
Basalt    & 23.592                 & 59.810              \\ \hline
Kimera    & 38.091                 & 68.327              \\ \hline
SVO2      & 6.463                  & 10.210              \\ \hline
\end{tabular}%
}
\caption{Average and peak CPU usage in \% of total available CPU, by the algorithms running the same experiment.}
\label{tab:cpu_usage}
\end{table}

\vspace{\baselineskip}

\noindent\emph{Computational resources:} From Table \ref{tab:cpu_usage} we can see that Lidar odometry algorithms LEGO LOAM and LOAM required the least CPU resources with 4.43\% and 6.63\% respectively due to their simpler algorithms. The CPU utilization of posegraph-based methods LIO SAM and HDL graph SLAM utilizing GTSAM and g2o is 21.4\% and 11.85\%, respectively, LIO SAM’s higher load reaching almost 80\% at peak can be explained by the additional IMU integration that it performs. The impact of feature detectors in the front-end can be observed from their higher average CPU usage compared to the rest in ORB SLAM3, Basalt VIO, and Kimera VIO. 

Among all the outdoor experiments, LIO SAM with top mount Lidar position was the most suitable for loader type machines used in our experiments, it had the least RMS of APE of 1.142m. Stability was more important to the IMU dependent LIO SAM whose performance degraded when the sensor was fixed to the loader bucket on the boom which has high vibrations while moving. LeGO LOAM was the most efficient while also having relatively good performance when the sensor is closer to the ground with an APE of 1.316m. With arbitrary LIDAR location, the simpler LOAM had reasonable results, hence making it a suitable choice when the sensor location is not fixed. Visual algorithms performed better when the camera was closer to the ground.

%\textcolor{blue}{Also some kind of conclusion, do those results correlate with the state-of-the-art results? What do those errors tell us? How do the results compare between indoor and outdoor? }

%\textcolor{blue}{We can also make a separate section of discussion, where to put those conclusions which I asked at the end of the outdoor and indoor subsubsections}

%\textcolor{red}{To be added:CPU usage, conclusion, related work, parameter table}
\section{CONCLUSION}
In this paper, we presented systematic evaluation of eight most popular state-of-the-art visual and Lidar SLAM methods. We tested them using our specially designed sensor suite which included visual sensors of different types, allowing us to capture the data from them simultaneously. The experiments were performed both outdoor and indoor, where we studied effects of sensor mounting position, terrain type, vibration effects, and variations in linear and angular velocities. 

Among the Lidar SLAM methods, LIO SLAM, using additionally IMU information, produces the smallest error in a single run. LEGO LOAM is suitable for lightweight applications but with sensor mounted closer to the ground. LOAM can used also in lightweight applications with less complex environments.
Comparing visual SLAM methods, ORB SLAM 3 performed well in dynamic and difficult environments. Basalt VIO deals better with quick changes in velocities. Thanks to the faster performing front-end, SVO could deal better with fast motion. Overall, Lidar's consistently were better with outdoor and with dynamic objects due to their large field of view of the environment.

\printbibliography[title=References]

\end{document}